\documentclass[conference]{IEEEtran}
\IEEEoverridecommandlockouts
\usepackage{cite}
\usepackage{amsmath,amssymb,amsfonts}
\usepackage{algorithmic}
\usepackage{graphicx}
\usepackage{textcomp}
\usepackage{xcolor}
\usepackage{booktabs}
\usepackage{tabularx}
\usepackage{multirow}
\usepackage{placeins}
\usepackage{url}
\usepackage[nolist]{acronym}
\usepackage{tikz}
\usepackage{pgfplots}
\pgfplotsset{compat=1.18}
\usepgfplotslibrary{statistics}

\def\BibTeX{{\rm B\kern-.05em{\sc i\kern-.025em b}\kern-.08em
    T\kern-.1667em\lower.7ex\hbox{E}\kern-.125emX}}
\begin{document}
\newacro{dso}[DSO]{distribution system operator}
\newacro{hd}[HD]{high-demand}
\newacro{lv}[LV]{low voltage}
\newacro{eu}[EU]{European Union}
\newacro{ml}[ML]{machine learning}
\newacro{sub}[SUB]{secondary substation}
\newacro{ac}[AC]{area code}
\newacro{fm}[FM]{foundation model}
\newacro{sb}[SB]{statistical baseline}
\newacro{stlf}[STLF]{short-term load forecasting}
\newacro{sm}[SM]{smart meter}
\newacro{ev}[EV]{electric vehicle}

\title{Peak-Aware Short-Term Load Forecasting Across Distribution Grid Aggregation Levels}

\author{
\IEEEauthorblockN{
Souhardya Chattopadhyay\textsuperscript{1,2\textdagger},
Julian Oelhaf\textsuperscript{1\textdagger}\textsuperscript{*},
Antonia Schoening\textsuperscript{2},
Jessica Deuschel\textsuperscript{2},
Bitan Bhattacharyya\textsuperscript{2},\\
Christian Bergler\textsuperscript{3},
Andreas Maier\textsuperscript{1},
Siming Bayer\textsuperscript{1}
}
\IEEEauthorblockA{
\textsuperscript{1}Pattern Recognition Lab, Friedrich-Alexander-Universität Erlangen-Nürnberg\\
\textsuperscript{2}Siemens AG, Smart Infrastructure\\
\textsuperscript{3}Department of Electrical Engineering, Media and Computer Science,
Ostbayerische Technische Hochschule Amberg-Weiden\\
Erlangen, Germany\\
\textsuperscript{*}Corresponding author: julian.oelhaf@fau.de
}
}

\maketitle
\begin{abstract}
For distribution system operators, short-term load forecasting (STLF) supports congestion management, voltage control, and asset protection. Most existing approaches focus on overall accuracy across all time steps and neglect performance during high-demand (HD) periods, where larger forecast errors can increase the risk of congestion and voltage violations. In this paper, we study peak-aware STLF across three operator-relevant distribution grid aggregation levels, area codes (AC), secondary substations (SUB), and low-voltage (LV) feeders, using open datasets from the United Kingdom and Switzerland. We compare statistical baselines, machine learning models (LightGBM and XGBoost), and recent time-series foundation models (Chronos-Bolt and Chronos-2) under a peak-aware evaluation framework that reports both overall and HD forecasting performance using NMAE and MAPE. The results show that Chronos-2 achieves the best HD performance across all aggregation levels, with HD-NMAE and HD-MAPE of 0.039 and 4.53\,\% at AC, 0.080 and 9.45\,\% at SUB, and 0.138 and 16.14\,\% at LV, while Chronos-Bolt consistently ranks second best. Compared with the gradient-boosted ML models, Chronos-2 reduces mean HD-NMAE by about 20--51\,\% across levels while remaining best or near-best on the overall metrics. A quantile analysis of the probabilistic Chronos outputs further identifies aggregation-specific operating points, and runtime measurements indicate that foundation-model inference is fast enough for practical deployment. Overall, the findings highlight peak-aware evaluation and aggregation-specific quantile selection as a practical pathway toward more operationally relevant STLF in distribution networks.
\end{abstract}


\setcounter{footnote}{0}
\renewcommand{\thefootnote}{\fnsymbol{footnote}}
\footnotetext{\textdagger These authors contributed equally to this work.}
\renewcommand{\thefootnote}{\arabic{footnote}}

\section{Introduction}

Energy systems are becoming harder to operate due to transport electrification, electric heating, distributed renewable generation, and rising demand variability at the grid edge. While these trends support decarbonization, they also increase short-term local fluctuations as electrified end uses such as \acp{ev}, heat pumps become more widespread and create sharper coincident peaks~\cite{uk_gov_ev_adoption,savanovic2024mitigating,toleikyte2024clean}. For \acp{dso}, this is especially critical at lower grid levels, where short demand surges can trigger congestion, voltage deviations, and asset stress~\cite{savanovic2024mitigating,toleikyte2024clean}. Hence, \ac{stlf} is both a planning task and an operational requirement for anticipating \ac{hd} situations early enough to support preventive action.

In practice, distribution forecasting is constrained by the level at which demand can be monitored and acted upon. Privacy and regulatory constraints often prevent the direct operational use of individual smart-meter readings, shifting attention to aggregated signals such as \ac{lv} feeders, \acp{sub}, and \ac{ac} demand~\cite{edps_metering_access_2022,knyrim2011smart,lee2021data}. Aggregation-aware forecasting is therefore essential, since signal behavior, operational relevance, and forecasting difficulty vary across aggregation levels.

A key difficulty is that standard evaluation is dominated by normal operating conditions. Since \ac{hd} intervals are typically sharper and more abrupt than normal demand patterns~\cite{savanovic2024mitigating}, models can achieve strong global accuracy while still making substantially larger errors around daily peaks~\cite{antoniadou2022effect}. Yet these are exactly the periods that matter most operationally, since peak-load misprediction can have disproportionate effects on cost and reliability~\cite{emde2023effects}. A model that performs well on average but poorly during \ac{hd} periods is therefore of limited value for distribution-grid operation.

\ac{stlf} is a mature field, and \ac{ml} methods have long improved over \acp{sb}. In particular, gradient-boosted trees such as XGBoost and LightGBM are widely used because they flexibly exploit calendar, weather, and lag features while remaining scalable and easy to train~\cite{uyar2025interpretable,abdalla2025machine,musbah2025forecasting,harikrishnan2024machine}. These studies show strong overall accuracy, but mostly under global or average metrics rather than explicit peak-aware evaluation~\cite{uyar2025interpretable,abdalla2025machine,musbah2025forecasting,harikrishnan2024machine}.

Recently, pretrained \acp{fm} have emerged as an alternative. Unlike supervised models trained for a specific dataset, they are pretrained on a large collection of time-series data and can be applied in an inference-driven manner. The Chronos family is a prominent example. Chronos introduced tokenization-based probabilistic forecasting, Chronos-Bolt emphasized faster inference~\cite{ansari2024chronos}, and Chronos-2~\cite{ansari2025chronos} extended the framework to a broader multivariate and covariate-aware setting. On electricity datasets such as fev-bench, Chronos-2 attains a 90.7\,\% win rate over other \acp{fm}~\cite{ansari2025chronos}, including TimesFM-2.5, Moirai-2.0, and Lag-Llama. These properties make Chronos models attractive for operational \ac{stlf}, especially when quantile outputs and long-context modeling are useful for peak-aware decision-making.

\definecolor{blue}{HTML}{0072B2}
\definecolor{red}{HTML}{A70107}

\begin{figure}[t]
\centering
\begin{tikzpicture}
\begin{axis}[
  width=9.4cm,
  height=4cm,
  ylabel={Scaled consumption},
  ymin=0.2, ymax=1.05,
  xmin=1, xmax=144,
  xtick={48,96,144},
  xticklabels={,,},
  xticklabel style={font=\fontsize{6.5}{5.5}\selectfont},
  yticklabel style={font=\fontsize{6.5}{5.5}\selectfont},
  unbounded coords=jump,
  label style={font=\fontsize{6.5}{5.5}\selectfont},
  legend style={
    font=\fontsize{6.5}{5.5}\selectfont,
    at={(0.5,1.02)},
    anchor=south,
    draw=none,
    fill=none,
    legend columns=3,
    /tikz/every even column/.append style={column sep=7pt},
  },
  clip=false,
]

\addplot+[
  thick,
  solid,
  mark=none,
  color=blue
] table[row sep=\\] {
x y\\
1 0.38\\ 2 0.36\\ 3 0.36\\ 4 0.31\\ 5 0.27\\ 6 0.26\\ 7 0.25\\ 8 0.26\\
9 0.23\\ 10 0.24\\ 11 0.31\\ 12 0.27\\ 13 0.29\\ 14 0.31\\ 15 0.31\\ 16 0.45\\
17 0.50\\ 18 0.46\\ 19 0.58\\ 20 0.44\\ 21 0.47\\ 22 0.66\\ 23 0.65\\ 24 0.69\\
25 0.67\\ 26 0.53\\ 27 0.52\\ 28 0.50\\ 29 0.57\\ 30 0.60\\ 31 0.58\\ 32 0.67\\
33 0.58\\ 34 0.71\\ 35 0.81\\ 36 0.90\\ 37 0.87\\ 38 0.93\\ 39 0.93\\ 40 1.00\\
41 0.91\\ 42 0.89\\ 43 0.80\\ 44 0.78\\ 45 0.80\\ 46 0.60\\ 47 0.58\\ 48 0.49\\
49 0.45\\ 50 0.44\\ 51 0.40\\ 52 0.40\\ 53 0.37\\ 54 0.32\\ 55 0.29\\ 56 0.26\\
57 0.26\\ 58 0.26\\ 59 0.27\\ 60 0.26\\ 61 0.30\\ 62 0.36\\ 63 0.35\\ 64 0.44\\
65 0.48\\ 66 0.48\\ 67 0.58\\ 68 0.66\\ 69 0.73\\ 70 0.78\\ 71 0.72\\ 72 0.76\\
73 0.78\\ 74 0.71\\ 75 0.80\\ 76 0.84\\ 77 0.76\\ 78 0.73\\ 79 0.76\\ 80 0.83\\
81 0.76\\ 82 0.81\\ 83 0.85\\ 84 0.99\\ 85 0.98\\ 86 0.93\\ 87 0.98\\ 88 0.91\\
89 0.91\\ 90 0.77\\ 91 0.82\\ 92 0.78\\ 93 0.66\\ 94 0.61\\ 95 0.59\\ 96 0.52\\
97 0.46\\ 98 0.44\\ 99 0.37\\ 100 0.34\\ 101 0.31\\ 102 0.29\\ 103 0.26\\ 104 0.27\\
105 0.26\\ 106 0.26\\ 107 0.28\\ 108 0.30\\ 109 0.30\\ 110 0.28\\ 111 0.33\\ 112 0.36\\
113 0.41\\ 114 0.49\\ 115 0.59\\ 116 0.70\\ 117 0.69\\ 118 0.82\\ 119 0.71\\ 120 0.75\\
121 0.80\\ 122 0.94\\ 123 0.97\\ 124 0.83\\ 125 0.87\\ 126 0.77\\ 127 0.81\\ 128 0.93\\
129 0.94\\ 130 0.80\\ 131 0.94\\ 132 1.00\\ 133 0.91\\ 134 0.88\\ 135 0.92\\ 136 0.93\\
137 0.73\\ 138 0.70\\ 139 0.79\\ 140 0.74\\ 141 0.70\\ 142 0.60\\ 143 0.47\\ 144 0.38\\
};
\addlegendentry{Scaled consumption}

\addplot+[
  thick,
  dashed,
  mark=none,
  color=red
] coordinates {(1,0.8) (144,0.8)};
\addlegendentry{HD-threshold}

\addplot[
  draw=red,
  draw opacity=0.1,
  fill=red,
  fill opacity=0.1,
  area legend
] coordinates {
  (35,0.2) (35,1.05) (43,1.05) (43,0.2) (35,0.2) (nan,nan)
  (75,0.2) (75,1.05) (76,1.05) (76,0.2) (75,0.2) (nan,nan)
  (82,0.2) (82,1.05) (89,1.05) (89,0.2) (82,0.2) (nan,nan)
  (121,0.2) (121,1.05) (125,1.05) (125,0.2) (121,0.2) (nan,nan)
  (127,0.2) (127,1.05) (136,1.05) (136,0.2) (127,0.2) (nan,nan)
};
\addlegendentry{HD-period}

\addplot[
  red,
  dotted,
  semithick,
  mark=none
] coordinates {
  (35,0.2) (35,1.05) (nan,nan)
  (43,0.2) (43,1.05) (nan,nan)
  (75,0.2) (75,1.05) (nan,nan)
  (76,0.2) (76,1.05) (nan,nan)
  (82,0.2) (82,1.05) (nan,nan)
  (89,0.2) (89,1.05) (nan,nan)
  (121,0.2) (121,1.05) (nan,nan)
  (125,0.2) (125,1.05) (nan,nan)
  (127,0.2) (127,1.05) (nan,nan)
  (136,0.2) (136,1.05) (nan,nan)
};
\addlegendentry{Start/End HD-Period}

\addplot+[
  only marks,
  mark=*,
  mark size=1pt,
  color=red,
  mark options={fill=red, draw=red},
  nodes near coords,
  nodes near coords align={above},
  every node near coord/.append style={
    font=\fontsize{6.5}{5.5}\selectfont,
    text=red,
    yshift=1pt
  },
  point meta=explicit symbolic
] coordinates {
  (45,0.80)
  (80,0.83)
  (91,0.82)
  (118,0.82)
};
\addlegendentry{Isolated HD points}

\node[anchor=north, font=\fontsize{6.5}{5.5}\selectfont] at (axis cs:24,0.2) {Day 1};
\node[anchor=north, font=\fontsize{6.5}{5.5}\selectfont] at (axis cs:72,0.2) {Day 2};
\node[anchor=north, font=\fontsize{6.5}{5.5}\selectfont] at (axis cs:120,0.2) {Day 3};

\end{axis}
\end{tikzpicture}
\caption[Three-day SUB demand profile with HD periods]{Typical three-day scaled consumption pattern at SUB level.}
\label{fig:SUB_scaled_consumption}
\end{figure}
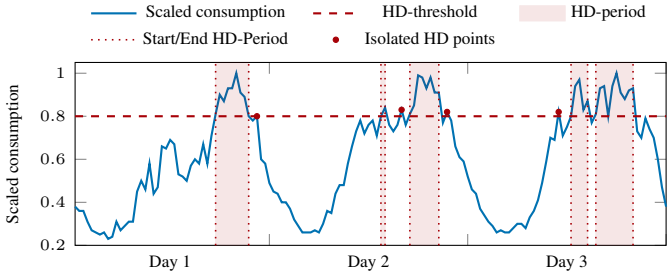


Despite substantial progress in \ac{stlf}, current evaluation practices remain misaligned with the requirements of distribution-grid operation. Most prior work reports performance averaged over all timestamps, which underrepresents \ac{hd} intervals where forecast errors are most critical for grid stability, congestion management, and asset utilization. In addition, existing studies often focus on a single aggregation level or a limited number of entities, rather than systematically evaluating \ac{dso}-relevant levels such as \ac{lv} feeders and \acp{sub}. Finally, while recent time-series \acp{fm} demonstrate strong general forecasting performance, their effectiveness under peak-aware evaluation and their behavior across aggregation levels, particularly with respect to probabilistic outputs, remain insufficiently understood.

This work addresses these limitations by investigating whether probabilistic \acp{fm} provide a measurable advantage over established statistical and \ac{ml} approaches for \ac{stlf} during operationally critical \ac{hd} intervals across distribution-grid aggregation levels. Specifically, this work makes the following contributions:
\begin{enumerate}
    \item[(i)] we introduce a peak-aware evaluation framework that explicitly separates \ac{hd} and non-\ac{hd} performance;
    \item[(ii)] we establish a scalable cross-aggregation benchmark across \ac{ac}, \ac{sub}, and \ac{lv} feeder levels with a large number of entities; and
    \item[(iii)] we analyze the operational behavior of probabilistic \acp{fm}, including aggregation-specific quantile selection and runtime considerations.
\end{enumerate}
By shifting the evaluation focus from average accuracy to peak-critical performance and by providing actionable guidance on model selection and operating points, this work supports more reliable \ac{stlf} and improved operational decision-making for \acp{dso}.

\section{Data and Forecasting Setup}
We study day-ahead \ac{stlf} across three aggregation levels that are relevant for distribution-grid monitoring and operation: \ac{ac}, \ac{sub}, and \ac{lv} feeder.

\subsection{Datasets and Aggregation Levels}
The \ac{ac}-level data are obtained from the open \ac{sm} dataset of the Swiss \ac{dso} CKW Group~\cite{axpo_smartmeter_dataset_b}, which provides aggregated consumption per postal-code region together with the number of contributing \acp{sm}. The \ac{sub} and \ac{lv} feeder datasets are taken from the Northern Powergrid open data portal in the United Kingdom~\cite{npg_aggregated_smartmeter}. At these three levels, an entity corresponds to one postal-code region, one substation, or one \ac{lv} feeder, respectively. All datasets cover the period from November~2023 to February~2025. A summary of the datasets used is provided in Table~\ref{tab:dataset_overview}.

\begin{table}[t]
\centering
\caption{Dataset overview and aggregation hierarchy across distribution grid levels.}
\label{tab:dataset_overview}
\footnotesize
\setlength{\tabcolsep}{3.5pt}
\renewcommand{\arraystretch}{1.10}
\begin{tabular}{c c c c}
\toprule
\textbf{Level} & \textbf{Aggregated from} & \textbf{Source} & \textbf{\# Entities} \\
\midrule
\ac{ac}  & Multiple \acp{sub}        & CKW Group (CH)~\cite{axpo_smartmeter_dataset_b} & 115 \\
\ac{sub} & Multiple \ac{lv} feeders  & Northern Powergrid (UK)~\cite{npg_aggregated_smartmeter} & 387 \\
\ac{lv}  & Multiple \acp{sm}         & Northern Powergrid (UK)~\cite{npg_aggregated_smartmeter} & 485 \\
\bottomrule
\end{tabular}
\end{table}

\subsection{Preprocessing and High-Demand Definition}
All time series are processed at the entity level. Duplicate timestamps are removed by aggregating consumption and \ac{sm} counts per entity and time step. Because the reported number of contributing \acp{sm} can vary due to communication or data-quality issues, we use per-meter demand $d_{e,t}$ as the primary target signal:
\begin{equation*}
d_{e,t} = \frac{E_{e,t}}{M_{e,t}},
\end{equation*}
where $E_{e,t}$ and $M_{e,t}$ denote aggregated energy consumption and the number of contributing \acp{sm} for entity $e$ at time $t$.

The CKW data are available at 15-minute resolution, whereas the Northern Powergrid data are provided at 30-minute resolution. To ensure comparability across aggregation levels, all series are aligned to a common 30-minute resolution by consecutive aggregation of CKW intervals.

To identify operationally critical periods in a scale-independent way, we use a rolling \ac{hd} normalization based on the recent history of each entity, given the lack of publicly available critical threshold data. For entity $e$ and forecast day $D$, let $s_{e,D}$ denote the 99th percentile ($p_{0.99}$) of per-meter demand observed during the preceding 14 days. Demand within day $D$ is then normalized as
\begin{equation*}
\tilde{d}_{e,t} = \frac{d_{e,t}}{\max(s_{e,D},\epsilon)}, \qquad t \in D,
\end{equation*}
where $\epsilon>0$ ensures numerical stability. A timestamp is classified as \ac{hd} if $\tilde{d}_{e,t}\geq 0.8$.
All remaining timestamps are treated as non-\ac{hd}. This definition allows a common peak-aware evaluation framework across aggregation levels without requiring explicit operational capacity limits for each entity. Figure~\ref{fig:SUB_scaled_consumption} shows a typical scaled consumption pattern at the \ac{sub} level together with the \ac{hd} threshold. Where such limits are known in practice, the same framework could be applied directly using those thresholds instead.

\subsection{Forecasting Task and Benchmark Setup}
For each entity, a 1-day-ahead forecast at 30-minute resolution is generated once per day at midnight, resulting in a horizon $H$ of 48 steps. The forecasting objective is evaluated over the period from March~1,~2024 to February~28,~2025 across \ac{ac}, \ac{sub}, and \ac{lv} feeder data following a rolling-history forecasting protocol. This setup enables a full-year comparison across seasons: Spring (March-May); Summer (June-August); Autumn (September-November); Winter (December-February), while preserving a sufficient look-back period for both \ac{ml} and foundation-model forecasting. This results in a large-scale benchmark comprising approximately 41{,}000 (\ac{ac}); 141{,}000 (\ac{sub}); and 177{,}000 (\ac{lv} feeders) daily forecasts, making it highly scalable and reliable for robust model evaluation across diverse grid conditions.

We use a compact set of exogenous covariates consisting of calendar features, holiday indicators, and weather variables. Holiday features are constructed using public holidays in England for Northern Powergrid entities and in Switzerland for CKW entities. Weather variables are obtained via Open-Meteo API~\cite{zippenfenig_openmeteo_2023} and include relative humidity (\%), feels-like temperature (°C), dew point temperature (°C), wind speed at 10 m (m/s), and global solar radiation (W/m²). The selection of these weather variables is guided by prior work demonstrating the relevance of meteorological factors for electricity consumption modeling~\cite{rahaman2024modeling,mosner2018seasonal}.

\section{Forecasting Methods and Evaluation}

We compare statistical baselines, conventional \ac{ml} models, and probabilistic foundation models.

\textbf{Statistical Baselines:} We include two same-time averaging \acp{sb} as simple references. One uses the mean at the same time of day over the previous 7 days, and the other over the previous 4 weeks. These baselines are computationally inexpensive and serve as low-complexity reference models.

\textbf{Machine Learning Models:} We evaluate LightGBM and XGBoost as representative gradient-boosted tree methods. Both use the feature set described above and produce point forecasts. Gradient boosting remains one of the most widely used and competitive paradigms in applied \ac{stlf}, combining flexible nonlinear learning with efficient training and robust handling of heterogeneous inputs~\cite{uyar2025interpretable,abdalla2025machine,musbah2025forecasting,harikrishnan2024machine}. This makes these models important references when testing whether newer \ac{fm} approaches offer additional benefit when evaluation shifts from average performance to \ac{hd} behavior.

\textbf{Foundation Models:} We further evaluate Chronos-Bolt and Chronos-2 as pretrained probabilistic time-series \acp{fm}~\cite{ansari2024chronos,ansari2025chronos}. Chronos-Bolt emphasizes efficient direct multi-step forecasting and is attractive from an inference-speed perspective. Chronos-2 extends the Chronos line toward broader multivariate and covariate-aware forecasting with longer context support. In contrast to the statistical baselines and standard boosted trees, both models provide quantile forecasts, which enables explicit analysis of conservative versus less conservative operating points under peak-aware evaluation.

\subsection{Training and Inference Protocol}
The forecasting setup follows a rolling-history protocol in which each next-day forecast uses only past information. For the \ac{ml} models, we use a fixed 90-day seasonal training window rather than daily per-entity retraining, which is computationally infeasible at this scale. As only limited history is available, the closest-matching prior seasonal block was selected using a small validation subset of separate entities from the same datasets that were not part of the final evaluation. This subset showed that warm months exhibit lower demand, whereas cold months show higher and more similar patterns, as illustrated in Figure~\ref{fig:daily_peak_by_season}. Accordingly, we forecast Spring~2024 using Winter~2023/24, Summer~2024 using Spring~2024, and both Autumn~2024 and Winter~2024/25 using Winter~2023/24. For the \acp{fm}, no retraining is required; instead, each test-day forecast is generated in a rolling manner from recent historical context only, with Chronos-Bolt using up to 2048 timestamps ($\sim$ 42 days) and Chronos-2 using the full 90-day context window under the chosen setup.

\definecolor{PR2}{HTML}{A8BCCD}
\definecolor{PR1}{HTML}{6BA2C0}
\definecolor{PR3}{HTML}{156082}
\definecolor{PR4}{HTML}{041E42}

\begin{figure}[t]
\centering
\begin{tikzpicture}
\pgfplotstableread[col sep=space]{
t autumn winter spring summer
0 327 nan nan nan
1 360 nan nan nan
2 386 nan nan nan
3 346 nan nan nan
4 367 nan nan nan
5 446 nan nan nan
6 348 nan nan nan
7 366 nan nan nan
8 346 nan nan nan
9 357 nan nan nan
10 431 nan nan nan
11 405 nan nan nan
12 384 nan nan nan
13 369 nan nan nan
14 379 nan nan nan
15 379 nan nan nan
16 365 nan nan nan
17 352 nan nan nan
18 401 nan nan nan
19 359 nan nan nan
20 366 nan nan nan
21 353 nan nan nan
22 365 nan nan nan
23 377 nan nan nan
24 405 nan nan nan
25 465 nan nan nan
26 370 nan nan nan
27 364 nan nan nan
28 375 nan nan nan
29 363 nan nan nan
30 nan 407 nan nan
31 nan 415 nan nan
32 nan 431 nan nan
33 nan 436 nan nan
34 nan 367 nan nan
35 nan 378 nan nan
36 nan 418 nan nan
37 nan 371 nan nan
38 nan 364 nan nan
39 nan 485 nan nan
40 nan 406 nan nan
41 nan 385 nan nan
42 nan 356 nan nan
43 nan 394 nan nan
44 nan 376 nan nan
45 nan 374 nan nan
46 nan 430 nan nan
47 nan 429 nan nan
48 nan 405 nan nan
49 nan 432 nan nan
50 nan 395 nan nan
51 nan 418 nan nan
52 nan 414 nan nan
53 nan 429 nan nan
54 nan 506 nan nan
55 nan 341 nan nan
56 nan 399 nan nan
57 nan 408 nan nan
58 nan 396 nan nan
59 nan 398 nan nan
60 nan 404 nan nan
61 nan 403 nan nan
62 nan 360 nan nan
63 nan 435 nan nan
64 nan 363 nan nan
65 nan 401 nan nan
66 nan 420 nan nan
67 nan 452 nan nan
68 nan 400 nan nan
69 nan 357 nan nan
70 nan 382 nan nan
71 nan 384 nan nan
72 nan 425 nan nan
73 nan 385 nan nan
74 nan 410 nan nan
75 nan 418 nan nan
76 nan 402 nan nan
77 nan 395 nan nan
78 nan 418 nan nan
79 nan 370 nan nan
80 nan 405 nan nan
81 nan 459 nan nan
82 nan 417 nan nan
83 nan 360 nan nan
84 nan 352 nan nan
85 nan 354 nan nan
86 nan 381 nan nan
87 nan 414 nan nan
88 nan 406 nan nan
89 nan 387 nan nan
90 nan 372 nan nan
91 nan 377 nan nan
92 nan 387 nan nan
93 nan 378 nan nan
94 nan 360 nan nan
95 nan 426 nan nan
96 nan 380 nan nan
97 nan 393 nan nan
98 nan 387 nan nan
99 nan 420 nan nan
100 nan 380 nan nan
101 nan 410 nan nan
102 nan 439 nan nan
103 nan 449 nan nan
104 nan 346 nan nan
105 nan 371 nan nan
106 nan 350 nan nan
107 nan 340 nan nan
108 nan 344 nan nan
109 nan 397 nan nan
110 nan 424 nan nan
111 nan 350 nan nan
112 nan 348 nan nan
113 nan 390 nan nan
114 nan 374 nan nan
115 nan 418 nan nan
116 nan 462 nan nan
117 nan 410 nan nan
118 nan 361 nan nan
119 nan 376 nan nan
120 nan 393 nan nan
121 nan nan 357 nan
122 nan nan 388 nan
123 nan nan 367 nan
124 nan nan 375 nan
125 nan nan 349 nan
126 nan nan 367 nan
127 nan nan 351 nan
128 nan nan 335 nan
129 nan nan 379 nan
130 nan nan 380 nan
131 nan nan 388 nan
132 nan nan 357 nan
133 nan nan 383 nan
134 nan nan 362 nan
135 nan nan 352 nan
136 nan nan 367 nan
137 nan nan 354 nan
138 nan nan 328 nan
139 nan nan 325 nan
140 nan nan 333 nan
141 nan nan 341 nan
142 nan nan 310 nan
143 nan nan 348 nan
144 nan nan 339 nan
145 nan nan 364 nan
146 nan nan 332 nan
147 nan nan 330 nan
148 nan nan 354 nan
149 nan nan 313 nan
150 nan nan 308 nan
151 nan nan 300 nan
152 nan nan 319 nan
153 nan nan 300 nan
154 nan nan 310 nan
155 nan nan 375 nan
156 nan nan 331 nan
157 nan nan 342 nan
158 nan nan 322 nan
159 nan nan 332 nan
160 nan nan 311 nan
161 nan nan 325 nan
162 nan nan 325 nan
163 nan nan 303 nan
164 nan nan 320 nan
165 nan nan 314 nan
166 nan nan 351 nan
167 nan nan 284 nan
168 nan nan 302 nan
169 nan nan 310 nan
170 nan nan 283 nan
171 nan nan 328 nan
172 nan nan 325 nan
173 nan nan 370 nan
174 nan nan 298 nan
175 nan nan 281 nan
176 nan nan 333 nan
177 nan nan 274 nan
178 nan nan 353 nan
179 nan nan 372 nan
180 nan nan 337 nan
181 nan nan 282 nan
182 nan nan 247 nan
183 nan nan 275 nan
184 nan nan 274 nan
185 nan nan 265 nan
186 nan nan 316 nan
187 nan nan 326 nan
188 nan nan 305 nan
189 nan nan 283 nan
190 nan nan 271 nan
191 nan nan 233 nan
192 nan nan 261 nan
193 nan nan 277 nan
194 nan nan 274 nan
195 nan nan 239 nan
196 nan nan 241 nan
197 nan nan 251 nan
198 nan nan 245 nan
199 nan nan 253 nan
200 nan nan 304 nan
201 nan nan 267 nan
202 nan nan 266 nan
203 nan nan 305 nan
204 nan nan 282 nan
205 nan nan 230 nan
206 nan nan 257 nan
207 nan nan 275 nan
208 nan nan 291 nan
209 nan nan 254 nan
210 nan nan 273 nan
211 nan nan 266 nan
212 nan nan 239 nan
213 nan nan nan 266
214 nan nan nan 309
215 nan nan nan 274
216 nan nan nan 273
217 nan nan nan 247
218 nan nan nan 270
219 nan nan nan 285
220 nan nan nan 297
221 nan nan nan 350
222 nan nan nan 329
223 nan nan nan 339
224 nan nan nan 301
225 nan nan nan 300
226 nan nan nan 299
227 nan nan nan 269
228 nan nan nan 325
229 nan nan nan 291
230 nan nan nan 299
231 nan nan nan 301
232 nan nan nan 303
233 nan nan nan 233
234 nan nan nan 231
235 nan nan nan 301
236 nan nan nan 282
237 nan nan nan 257
238 nan nan nan 266
239 nan nan nan 252
240 nan nan nan 286
241 nan nan nan 276
242 nan nan nan 301
243 nan nan nan 267
244 nan nan nan 251
245 nan nan nan 255
246 nan nan nan 266
247 nan nan nan 254
248 nan nan nan 264
249 nan nan nan 328
250 nan nan nan 282
251 nan nan nan 278
252 nan nan nan 274
253 nan nan nan 278
254 nan nan nan 290
255 nan nan nan 349
256 nan nan nan 338
257 nan nan nan 274
258 nan nan nan 281
259 nan nan nan 243
260 nan nan nan 272
261 nan nan nan 223
262 nan nan nan 262
263 nan nan nan 237
264 nan nan nan 261
265 nan nan nan 240
266 nan nan nan 261
267 nan nan nan 298
268 nan nan nan 236
269 nan nan nan 231
270 nan nan nan 254
271 nan nan nan 280
272 nan nan nan 218
273 nan nan nan 201
274 nan nan nan 287
275 nan nan nan 235
276 nan nan nan 237
277 nan nan nan 245
278 nan nan nan 247
279 nan nan nan 241
280 nan nan nan 258
281 nan nan nan 270
282 nan nan nan 269
283 nan nan nan 246
284 nan nan nan 263
285 nan nan nan 242
286 nan nan nan 230
287 nan nan nan 258
288 nan nan nan 259
289 nan nan nan 255
290 nan nan nan 245
291 nan nan nan 291
292 nan nan nan 272
293 nan nan nan 259
294 nan nan nan 231
295 nan nan nan 302
296 nan nan nan 249
297 nan nan nan 245
298 nan nan nan 252
299 nan nan nan 297
300 nan nan nan 324
301 nan nan nan 306
302 nan nan nan 233
303 nan nan nan 261
304 nan nan nan 229
305 340 nan nan nan
306 347 nan nan nan
307 256 nan nan nan
308 262 nan nan nan
309 324 nan nan nan
310 300 nan nan nan
311 277 nan nan nan
312 337 nan nan nan
313 298 nan nan nan
314 281 nan nan nan
315 297 nan nan nan
316 286 nan nan nan
317 288 nan nan nan
318 309 nan nan nan
319 309 nan nan nan
320 266 nan nan nan
321 240 nan nan nan
322 280 nan nan nan
323 301 nan nan nan
324 310 nan nan nan
325 313 nan nan nan
326 346 nan nan nan
327 318 nan nan nan
328 291 nan nan nan
329 324 nan nan nan
330 329 nan nan nan
331 312 nan nan nan
332 361 nan nan nan
333 375 nan nan nan
334 358 nan nan nan
335 319 nan nan nan
336 331 nan nan nan
337 326 nan nan nan
338 308 nan nan nan
339 349 nan nan nan
340 410 nan nan nan
341 325 nan nan nan
342 305 nan nan nan
343 359 nan nan nan
344 331 nan nan nan
345 313 nan nan nan
346 390 nan nan nan
347 385 nan nan nan
348 354 nan nan nan
349 318 nan nan nan
350 389 nan nan nan
351 312 nan nan nan
352 351 nan nan nan
353 319 nan nan nan
354 400 nan nan nan
355 315 nan nan nan
356 298 nan nan nan
357 343 nan nan nan
358 355 nan nan nan
359 328 nan nan nan
360 341 nan nan nan
361 367 nan nan nan
362 358 nan nan nan
363 300 nan nan nan
364 331 nan nan nan
365 371 nan nan nan
366 303 nan nan nan
367 332 nan nan nan
368 372 nan nan nan
369 383 nan nan nan
370 363 nan nan nan
371 344 nan nan nan
372 335 nan nan nan
373 351 nan nan nan
374 327 nan nan nan
375 433 nan nan nan
376 383 nan nan nan
377 344 nan nan nan
378 366 nan nan nan
379 366 nan nan nan
380 333 nan nan nan
381 332 nan nan nan
382 452 nan nan nan
383 425 nan nan nan
384 403 nan nan nan
385 393 nan nan nan
386 402 nan nan nan
387 347 nan nan nan
388 401 nan nan nan
389 467 nan nan nan
390 383 nan nan nan
391 346 nan nan nan
392 390 nan nan nan
393 378 nan nan nan
394 349 nan nan nan
395 nan 398 nan nan
396 nan 438 nan nan
397 nan 393 nan nan
398 nan 382 nan nan
399 nan 386 nan nan
400 nan 349 nan nan
401 nan 352 nan nan
402 nan 360 nan nan
403 nan 448 nan nan
404 nan 384 nan nan
405 nan 370 nan nan
406 nan 379 nan nan
407 nan 359 nan nan
408 nan 331 nan nan
409 nan 356 nan nan
410 nan 450 nan nan
411 nan 431 nan nan
412 nan 366 nan nan
413 nan 355 nan nan
414 nan 370 nan nan
415 nan 391 nan nan
416 nan 422 nan nan
417 nan 491 nan nan
418 nan 421 nan nan
419 nan 426 nan nan
420 nan 499 nan nan
421 nan 413 nan nan
422 nan 395 nan nan
423 nan 417 nan nan
424 nan 413 nan nan
425 nan 438 nan nan
426 nan 387 nan nan
427 nan 390 nan nan
428 nan 479 nan nan
429 nan 448 nan nan
430 nan 411 nan nan
431 nan 486 nan nan
432 nan 425 nan nan
433 nan 410 nan nan
434 nan 422 nan nan
435 nan 437 nan nan
436 nan 399 nan nan
437 nan 441 nan nan
438 nan 532 nan nan
439 nan 404 nan nan
440 nan 366 nan nan
441 nan 350 nan nan
442 nan 385 nan nan
443 nan 409 nan nan
444 nan 386 nan nan
445 nan 508 nan nan
446 nan 408 nan nan
447 nan 357 nan nan
448 nan 379 nan nan
449 nan 399 nan nan
450 nan 410 nan nan
451 nan 364 nan nan
452 nan 533 nan nan
453 nan 400 nan nan
454 nan 395 nan nan
455 nan 355 nan nan
456 nan 448 nan nan
457 nan 421 nan nan
458 nan 370 nan nan
459 nan 470 nan nan
460 nan 399 nan nan
461 nan 317 nan nan
462 nan 375 nan nan
463 nan 385 nan nan
464 nan 397 nan nan
465 nan 394 nan nan
466 nan 467 nan nan
467 nan 398 nan nan
468 nan 358 nan nan
469 nan 392 nan nan
470 nan 431 nan nan
471 nan 462 nan nan
472 nan 403 nan nan
473 nan 512 nan nan
474 nan 416 nan nan
475 nan 335 nan nan
476 nan 358 nan nan
477 nan 396 nan nan
478 nan 363 nan nan
479 nan 339 nan nan
480 nan 376 nan nan
481 nan 380 nan nan
482 nan 304 nan nan
483 nan 352 nan nan
484 nan 334 nan nan
485 nan 355 nan nan
}\datatable

\begin{axis}[
  width=\columnwidth,
  height=5.2cm,
  ylabel={Peak consumption (Wh)},
  xmin=0, xmax=485,
  ymin=180, ymax=560,
  unbounded coords=jump,
  xtick={30,121,213,305,396},
  xticklabels={Dec 23,Mar 24,Jun 24,Sep 24,Dec 24},
  x tick label style={font=\fontsize{6.5}{5.5}\selectfont, rotate=0, anchor=north},
  y tick label style={font=\fontsize{6.5}{5.5}\selectfont},
  label style={font=\fontsize{6.5}{5.5}\selectfont},
  legend style={
    font=\fontsize{6.5}{5.5}\selectfont,
    at={(0.5,1.02)},
    anchor=south,
    draw=none,
    legend columns=4,
    /tikz/every even column/.append style={column sep=6pt},
  },
  tick label style={/pgf/number format/fixed},
  clip=false
]

\addplot[PR4, solid, thick] table[x=t,y=autumn] {\datatable};
\addlegendentry{Autumn}

\addplot[PR1, solid, thick] table[x=t,y=winter] {\datatable};
\addlegendentry{Winter}

\addplot[PR2, solid, thick] table[x=t,y=spring] {\datatable};
\addlegendentry{Spring}

\addplot[PR3, solid, thick] table[x=t,y=summer] {\datatable};
\addlegendentry{Summer}

\end{axis}
\end{tikzpicture}
\caption{Daily peak consumption pattern at \ac{sub} level.}
\label{fig:daily_peak_by_season}
\end{figure}
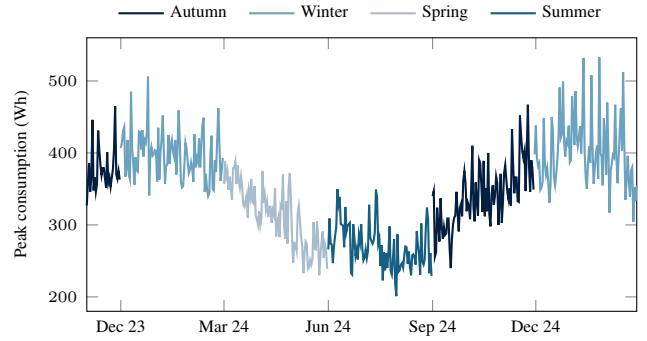

\begin{table*}[ht]
\centering
\caption{NMAE \& MAPE performance (mean $\pm$ std) across aggregation levels for all models. For each aggregation level and metric column, the lowest mean is shown in \textbf{bold} and the second-lowest mean is \underline{underlined}.}
\label{tab:main_results}
\footnotesize
\setlength{\tabcolsep}{6.2pt}
\renewcommand{\arraystretch}{1.0}
\begin{tabular}{c l l c c c c c c}
\toprule
\textbf{Level} & \textbf{Type} & \textbf{Model} & \textbf{NMAE} & \textbf{HD-NMAE} & \textbf{Non HD-NMAE} & \textbf{MAPE} & \textbf{HD-MAPE} & \textbf{Non HD-MAPE} \\
\midrule
\multirow{6}{*}{\ac{ac}}
& \ac{sb} & Last 7-day avg   & 0.055 $\pm$ 0.018 & 0.059 $\pm$ 0.021 & 0.054 $\pm$ 0.017 & 9.75 $\pm$ 3.15\,\% & 6.84 $\pm$ 2.44\,\% & 10.18 $\pm$ 3.32\,\% \\
& \ac{sb} & Last 4-week avg  & 0.066 $\pm$ 0.027 & 0.073 $\pm$ 0.032 & 0.064 $\pm$ 0.026 & 11.54 $\pm$ 4.79\,\% & 8.34 $\pm$ 3.61\,\% & 11.96 $\pm$ 4.95\,\% \\
& \ac{ml} & LightGBM         & 0.080 $\pm$ 0.032 & 0.080 $\pm$ 0.027 & 0.080 $\pm$ 0.033 & 14.65 $\pm$ 7.41\,\% & 9.11 $\pm$ 3.14\,\% & 15.49 $\pm$ 7.76\,\% \\
& \ac{ml} & XGBoost          & 0.081 $\pm$ 0.033 & 0.080 $\pm$ 0.027 & 0.082 $\pm$ 0.034 & 14.98 $\pm$ 7.57\,\% & 9.10 $\pm$ 3.11\,\% & 15.87 $\pm$ 7.93\,\% \\
& \ac{fm} & Chronos-Bolt     & \underline{0.051} $\pm$ 0.007 & \underline{0.053} $\pm$ 0.010 & \underline{0.051} $\pm$ 0.008 & \underline{9.21} $\pm$ 1.57\,\% & \underline{6.11} $\pm$ 1.12\,\% & \underline{9.78} $\pm$ 1.65\,\% \\
& \ac{fm} & Chronos-2         & \textbf{0.042} $\pm$ 0.007 & \textbf{0.039} $\pm$ 0.007 & \textbf{0.043} $\pm$ 0.008 & \textbf{7.68} $\pm$ 1.45\,\% & \textbf{4.53} $\pm$ 0.89\,\% & \textbf{8.25} $\pm$ 1.56\,\% \\
\midrule
\multirow{6}{*}{\ac{sub}}
& \ac{sb} & Last 7-day avg   & \textbf{0.074} $\pm$ 0.012 & 0.117 $\pm$ 0.034 & \textbf{0.068} $\pm$ 0.008 & \textbf{14.96} $\pm$ 2.21\,\% & 13.54 $\pm$ 3.88\,\% & \textbf{15.01} $\pm$ 2.21\,\% \\
& \ac{sb} & Last 4-week avg  & 0.078 $\pm$ 0.011 & 0.119 $\pm$ 0.031 & \underline{0.072} $\pm$ 0.010 & \underline{15.78} $\pm$ 2.52\,\% & 13.80 $\pm$ 3.53\,\% & \underline{15.98} $\pm$ 2.69\,\% \\
& \ac{ml} & LightGBM         & 0.098 $\pm$ 0.021 & 0.113 $\pm$ 0.024 & 0.096 $\pm$ 0.023 & 20.21 $\pm$ 4.62\,\% & 13.19 $\pm$ 2.74\,\% & 21.08 $\pm$ 4.93\,\% \\
& \ac{ml} & XGBoost          & 0.098 $\pm$ 0.022 & 0.113 $\pm$ 0.024 & 0.096 $\pm$ 0.023 & 20.42 $\pm$ 4.75\,\% & 13.20 $\pm$ 2.69\,\% & 21.31 $\pm$ 5.09\,\% \\
& \ac{fm} & Chronos-Bolt     & \underline{0.075} $\pm$ 0.010 & \underline{0.096} $\pm$ 0.021 & 0.073 $\pm$ 0.009 & \underline{15.78} $\pm$ 2.30\,\% & \underline{11.29} $\pm$ 2.43\,\% & 16.35 $\pm$ 2.42\,\% \\
& \ac{fm} & Chronos-2         & \textbf{0.074} $\pm$ 0.009 & \textbf{0.080} $\pm$ 0.019 & 0.073 $\pm$ 0.008 & 15.81 $\pm$ 2.25\,\% & \textbf{9.45} $\pm$ 2.21\,\% & 16.61 $\pm$ 2.36\,\% \\
\midrule
\multirow{6}{*}{\ac{lv}}
& \ac{sb} & Last 7-day avg   & \textbf{0.091} $\pm$ 0.010 & 0.199 $\pm$ 0.039 & \textbf{0.082} $\pm$ 0.007 & \textbf{20.64} $\pm$ 2.42\,\% & 22.81 $\pm$ 4.37\,\% & \textbf{20.41} $\pm$ 2.51\,\% \\
& \ac{sb} & Last 4-week avg  & 0.096 $\pm$ 0.010 & 0.197 $\pm$ 0.037 & \underline{0.088} $\pm$ 0.009 & \underline{21.86} $\pm$ 2.82\,\% & 22.67 $\pm$ 4.16\,\% & \underline{21.80} $\pm$ 2.99\,\% \\
& \ac{ml} & LightGBM         & 0.115 $\pm$ 0.018 & 0.173 $\pm$ 0.031 & 0.110 $\pm$ 0.019 & 26.96 $\pm$ 4.81\,\% & 20.07 $\pm$ 3.44\,\% & 27.52 $\pm$ 5.17\,\% \\
& \ac{ml} & XGBoost          & 0.117 $\pm$ 0.019 & 0.173 $\pm$ 0.031 & 0.112 $\pm$ 0.021 & 27.49 $\pm$ 5.21\,\% & 20.02 $\pm$ 3.40\,\% & 28.10 $\pm$ 5.59\,\% \\
& \ac{fm} & Chronos-Bolt     & \underline{0.094} $\pm$ 0.010 & \underline{0.160} $\pm$ 0.034 & 0.089 $\pm$ 0.009 & 22.14 $\pm$ 2.76\,\% & \underline{18.62} $\pm$ 3.75\,\% & 22.47 $\pm$ 2.88\,\% \\
& \ac{fm} & Chronos-2         & \underline{0.094} $\pm$ 0.009 & \textbf{0.138} $\pm$ 0.030 & 0.091 $\pm$ 0.008 & 22.80 $\pm$ 2.77\,\% & \textbf{16.14} $\pm$ 3.40\,\% & 23.36 $\pm$ 2.84\,\% \\
\bottomrule
\end{tabular}
\end{table*}

\subsection{Peak-Aware Evaluation Metric}
To evaluate both overall forecasting quality and performance during operationally critical periods, all metrics are computed over three timestamp subsets: 

(i) all points: $\mathcal{T}$, 

(ii) \ac{hd} points: $\mathcal{T}_{\mathrm{HD}}=\{t\in\mathcal{T}\mid \tilde{y}_t\geq 0.8\}$, and 

(iii) non-\ac{hd} points: $\mathcal{T}_{\mathrm{nonHD}}=\{t\in\mathcal{T}\mid \tilde{y}_t<0.8\}$.

Here, $y_t$ and $\hat{y}_t$ denote the actual and forecast demand, and $\tilde{y}_t$ and $\hat{\tilde{y}}_t$ denote their normalized counterparts. For any subset $\mathcal{S}\in\{\mathcal{T},\mathcal{T}_{\mathrm{HD}},\mathcal{T}_{\mathrm{non-HD}}\}$, we define the normalized mean absolute error (NMAE) and the mean absolute percentage error (MAPE) as
\begin{equation*}
\mathrm{NMAE}(\mathcal{S})=
\frac{1}{|\mathcal{S}|}\sum_{t\in\mathcal{S}}
\left|\tilde{y}_t-\hat{\tilde{y}}_t\right|
\end{equation*}
\begin{equation*}
\mathrm{MAPE}(\mathcal{S})=
\frac{100}{|\mathcal{S}|}\sum_{t\in\mathcal{S}}
\left|\frac{y_t-\hat{y}_t}{y_t}\right|
\end{equation*}

Thus, overall, HD, and non-HD metrics are obtained by setting $\mathcal{S}=\mathcal{T}$, $\mathcal{T}_{\mathrm{HD}}$, and $\mathcal{T}_{\mathrm{nonHD}}$, respectively. All reported values are computed day-wise and then aggregated over the evaluation period.

\subsection{Quantile Selection for Probabilistic Forecasting}
Chronos-Bolt and Chronos-2 produce multiple quantile forecasts rather than a single point prediction. Since the operating quantile controls the degree of conservativeness, we do not assume that the median forecast is automatically optimal. Instead, we evaluate 11 quantiles, $q_{0.50}, q_{0.55}, \dots, q_{0.95}, q_{0.99}$, using the same overall, \ac{hd}, and non-\ac{hd} metrics described above. Quantile selection is then performed on a small validation set by jointly considering overall competitiveness and peak-aware performance. The selected quantiles are used in the final benchmark reported in the Results section.

\section{Results}

\subsection{Performance Across Aggregation Levels}

Table~\ref{tab:main_results} summarizes forecasting performance across models and aggregation levels. Chronos-Bolt and Chronos-2 are reported using the aggregation-specific operating quantiles selected on the validation set (Table~\ref{tab:best_quantiles}); their behavior is analyzed in more detail in Section~\ref{sec:quantile_analysis}. Across all levels, the Chronos models dominate the peak-critical metrics. Chronos-2 achieves the best HD-NMAE and HD-MAPE in every setting, and Chronos-Bolt is consistently second-best. This indicates that the \acp{fm} are more reliable for capturing peak (HD) intervals while maintaining overall performance that is competitive with the best traditional models. At \ac{ac} level, this advantage extends to the full evaluation, where Chronos-2 is best on both overall and \ac{hd} metrics, improving over XGBoost by 48.1\,\% in overall NMAE and 50.2\,\% in HD-MAPE.

At \ac{sub} and \ac{lv} levels, the Chronos models remain competitive overall, but their clearest advantage appears in the \ac{hd} performance. At \ac{lv} level, for example, Chronos-2 is only 3.3\,\% worse in overall NMAE than the best competing model, yet it improves HD-NMAE by 30.7\,\% and HD-MAPE by 29.2\,\%. This pattern is operationally important because lower aggregation levels exhibit sharper and less regular peaks, making accurate \ac{hd} forecasting substantially more difficult for both baselines and boosted tree models.

The entity-level extremes denoting the best and worst performing entities further confirm the robustness of the Chronos models. Chronos-2 achieves both the lowest best-case and lowest worst-case HD-MAPE across entities, indicating superior consistency and robustness compared to baselines. Its best-case HD-MAPE reaches 2.47\,\% at \ac{ac}, 5.48\,\% at \ac{sub}, and 6.91\,\% at \ac{lv} feeder. The robustness is especially evident at \ac{sub} and \ac{lv} feeder levels, where the worst-case HD-MAPE is 39.54\,\% and 45.67\,\%, respectively, compared with values above 48\,\% and 51\,\% for the \acp{sb} and \ac{ml} models.

Seasonal results show the same pattern. At \ac{ac} and \ac{sub} levels, Summer and Spring are the most difficult seasons, respectively, yielding the highest HD-MAPE across model families. Chronos-2 nevertheless shows a much smaller gap between best and worst seasonal HD-MAPE, indicating more stable performance across the year. Its seasonal spread is only 0.68\,\% points at \ac{ac} and 1.86\,\% points at \ac{sub}, compared with about 2--3.5\,\% points for the \acp{sb} and \ac{ml} models.

\subsection{Quantile Behavior}\label{sec:quantile_analysis}

Quantile selection was performed on a validation set separately for each aggregation level and Chronos model. The rule was: (i) compute, for each candidate quantile, the overall NMAE difference to the best non-Chronos model at the same level and rank candidates in ascending order; (ii) retain the top overall candidates before the HD-NMAE begins to rise again, since HD-NMAE typically decreases as overall NMAE increases up to a turning point; and (iii) among these retained candidates, select the quantile with the lowest HD-NMAE and HD-MAPE. If multiple candidates remained, the one with the smaller overall NMAE gap was chosen. A slightly worse overall candidate was selected only if it improved both \ac{hd} metrics relative to the better retained candidate.

This rule yielded level-dependent operating quantiles, reported in Table~\ref{tab:best_quantiles}. At the \ac{sub} level, for example, Chronos-2 $q_{0.65}$ and $q_{0.70}$ were the two retained candidates because they had the smallest overall NMAE gap to the best non-Chronos model. Here, $q_{0.65}$ improved on the best non-Chronos model by 6.8\,\%, whereas $q_{0.70}$ matched it exactly. Since $q_{0.70}$ further reduced HD-NMAE by 7.0\,\% relative to $q_{0.65}$, it was selected. The same selection logic was applied at other levels.

\begin{table}[ht]
\centering
\caption{Best quantile per model and aggregation level.}
\label{tab:best_quantiles}
\footnotesize
\setlength{\tabcolsep}{7pt}
\renewcommand{\arraystretch}{1}
\begin{tabular}{l c c}
\toprule
\textbf{Level} & \textbf{Chronos-Bolt} & \textbf{Chronos-2} \\
\midrule
\ac{ac}       & $q_{0.65}$ & $q_{0.65}$ \\
\ac{sub}       & $q_{0.65}$ & $q_{0.70}$ \\
\ac{lv}         & $q_{0.65}$ & $q_{0.70}$ \\
\bottomrule
\end{tabular}
\end{table}

\subsection{Runtime and Deployment}

We report median inference latency per 24-hour forecast for Chronos-Bolt and Chronos-2 on CPU and an NVIDIA A100 GPU. Median CPU and GPU inference times are 1.3\,s and 0.007\,s (Chronos-Bolt) and 9.8\,s and 0.2\,s (Chronos-2), respectively, resulting in a 185$\times$ speedup for Chronos-Bolt and 49$\times$ speedup for Chronos-2 on GPU relative to CPU. For deployment, Chronos-2 provides the strongest accuracy–robustness trade-off, while Chronos-Bolt offers competitive accuracy at substantially lower latency.

\section{Conclusion}

This paper investigated peak-aware \acl{stlf} across \ac{ac}, \ac{sub}, and \ac{lv} aggregation levels. Chronos-based \acp{fm} consistently outperform established baselines on the operationally critical \ac{hd} intervals while maintaining competitive overall accuracy.
Chronos-2 delivers the strongest performance, reducing \ac{hd} errors by around 30\% relative to the best non-Chronos models without sacrificing overall accuracy. This advantage becomes more pronounced at lower aggregation levels, where load profiles are more volatile and peak behavior is harder to predict.
With GPU inference times of 0.007\,s for Chronos-Bolt and 0.2\,s for Chronos-2 per 24-hour forecast, both models are directly applicable in operational settings. Overall, \acp{fm} provide a clear and practical advantage for peak-aware \ac{stlf}, enabling more reliable monitoring of critical demand periods in distribution grids.
Future work will integrate Chronos-2 forecasts into model predictive control to enable uncertainty-aware demand management and grid-stability optimization under real-time operational constraints.

\section*{Acknowledgment}

This project was funded by the Deutsche Forschungsgemeinschaft (DFG, German Research Foundation) - 535389056.
\vfill
\pagebreak

\bibliographystyle{IEEEtran}
\bibliography{references}

@misc{edps_metering_access_2022,
  author       = {{European Data Protection Supervisor (EDPS)}},
  title        = {{EDPS Formal comments on the draft Commission Implementing Regulation on interoperability requirements and non-discriminatory and transparent procedures for access to metering and consumption data}},
  year         = {2022},
  howpublished = {\url{https://www.edps.europa.eu/system/files/2022-09/22-08-24_access-metering-and-consumption-data_en.pdf}},
  note         = {Accessed: Nov. 1, 2025}
}

@misc{knyrim2011smart,
  title={{Smart metering under EU Data Protection Law. International Data Privacy Law}},
  author={Knyrim, Rainer and Trieb, Gerald},
  year={2011},
  publisher={Oxford. First published online March}
}

@article{lee2021data,
  title={{Data privacy and residential smart meters: Comparative analysis and harmonization potential}},
  author={Lee, Dasom and Hess, David J},
  journal={Utilities Policy},
  volume={70},
  pages={101188},
  year={2021},
  publisher={Elsevier}
}

@misc{axpo_smartmeter_dataset_b,
  author       = {{CKW AG}},
  title        = {{CKW Open Data Smart Meter: Dataset B - Aggregated smart meter data}},
  year         = {2025},
  howpublished = {\url{https://open.data.axpo.com/}},
  note         = {Accessed: Dec. 6, 2025.}
}

@misc{npg_aggregated_smartmeter,
  author       = {{Northern Powergrid}},
  title        = {Aggregated Smart Metering Dataset},
  year         = {2025},
  howpublished = {\url{https://northernpowergrid.opendatasoft.com/}},
  note         = {Accessed: Nov. 04, 2025.}
}

@misc{zippenfenig_openmeteo_2023,
  author       = {Zippenfenig, Patrick},
  title        = {Open-Meteo.com Weather API},
  year         = {2023},
  publisher    = {Zenodo},
  doi          = {10.5281/zenodo.7970649},
  url          = {https://open-meteo.com/},
  note         = {Accessed 2025-12-23}
}

@article{uyar2025interpretable,
  title={{Interpretable building energy performance prediction using XGBoost Quantile Regression}},
  author={Uyar, Sinem Guler Kangalli and Ozbay, Bilge Kagan and Dal, Berker},
  journal={Energy and Buildings},
  volume={340},
  pages={115815},
  year={2025},
  publisher={Elsevier}
}

@article{abdalla2025machine,
  title={{Machine learning-based residential load demand forecasting: Evaluating ELM, XGBoost, RF, and SVM for enhanced energy system and sustainability}},
  author={Abdalla, Modawy Adam Ali and Ishaga, Ahmed Mohamed and Osman, Hassan Ahmed and Elhindi, Mohamed and Ibrahim, Nasreldin and Snani, Aissa and Hamid, Gomaa Haroun Ali and Hammad, Abdallah},
  journal={Science in Information Technology Letters},
  volume={6},
  number={1},
  pages={1--15},
  year={2025}
}

@article{musbah2025forecasting,
  title={Forecasting load consumption: a comprehensive evaluation of deep learning and machine learning techniques},
  author={Musbah, Hmeda and Elsaraiti, Meftah},
  journal={Electric Power Systems Research},
  volume={247},
  pages={111834},
  year={2025},
  publisher={Elsevier}
}

@inproceedings{harikrishnan2024machine,
  title={{Machine Learning Approaches for Load Forecasting and Time Series Analysis}},
  author={Harikrishnan, GR and Premnath, Tarun and Pranav, S and Varghese, Sandra Mariya and Krishna, Sandra and Sreedharan, Sasidharan},
  booktitle={2024 7th International Conference on Circuit Power and Computing Technologies (ICCPCT)},
  volume={1},
  pages={1739--1745},
  year={2024},
  organization={IEEE}
}

@article{ansari2024chronos,
  title={Chronos: Learning the language of time series},
  author={Ansari, Abdul Fatir and Stella, Lorenzo and Turkmen, Caner and Zhang, Xiyuan and Mercado, Pedro and Shen, Huibin and Shchur, Oleksandr and Rangapuram, Syama Sundar and Arango, Sebastian Pineda and Kapoor, Shubham and others},
  journal={arXiv preprint arXiv:2403.07815},
  year={2024}
}

@article{ansari2025chronos,
  title={Chronos-2: From univariate to universal forecasting},
  author={Ansari, Abdul Fatir and Shchur, Oleksandr and K{\"u}ken, Jaris and Auer, Andreas and Han, Boran and Mercado, Pedro and Rangapuram, Syama Sundar and Shen, Huibin and Stella, Lorenzo and Zhang, Xiyuan and others},
  journal={arXiv preprint arXiv:2510.15821},
  year={2025}
}

@article{rahaman2024modeling,
  title={Modeling influence of weather variables on energy consumption in an agricultural research institute in Ibadan, Nigeria},
  author={Rahaman, Abu and Amakor, John and Kazeem, Rasaq and Olugasa, Temilola and Ajide, Olusegun and Idusuyi, Nosa and Jen, Tien-Chien and Akinlabi, Esther},
  journal={AIMS energy},
  volume={12},
  number={1},
  pages={256--270},
  year={2024},
  publisher={Amer Inst Mathematical Sciences-Aims}
}

@article{mosner2018seasonal,
  title={The Seasonal Effects of Weather on Residential Electric-Energy Usage},
  author={Mosner-Ansong, KF and Duah, D},
  journal={Journal of Energy and Natural Resource Management},
  volume={1},
  number={1},
  year={2018}
}

@inproceedings{antoniadou2022effect,
  title={Effect of short-term and high-resolution load forecasting errors on microgrid operation costs},
  author={Antoniadou-Plytaria, Kyriaki and Eriksson, Ludvig and Johansson, Jakob and Johnsson, Richard and K{\"o}tz, Lasse and Lamm, Johan and Lundblad, Ellinor and Steen, David and Tuan, Le Anh and Carlson, Ola},
  booktitle={2022 IEEE PES Innovative Smart Grid Technologies Conference Europe (ISGT-Europe)},
  pages={1--5},
  year={2022},
  organization={IEEE}
}

@article{emde2023effects,
  title={Effects of load forecast deviation on the specification of energy storage systems},
  author={Emde, Alexander and M{\"a}rkle, Lisa and Kratzer, Benedikt and Schnell, Felix and Baur, Lukas and Sauer, Alexander},
  journal={Designs},
  volume={7},
  number={5},
  pages={107},
  year={2023},
  publisher={MDPI}
}

@misc{uk_gov_ev_adoption,
  author       = {{Department of Transport, UK Gov}},
  title        = {Vehicle licensing statistics: 2024},
  year         = {2024},
  howpublished = {\url{https://www.gov.uk/government/statistics/vehicle-licensing-statistics-2024}},
  note         = {Contains public sector information licensed under the Open Government Licence v3.0.}
}

@article{savanovic2024mitigating,
  title={Mitigating the charging rush hour},
  author={Savanovic, Milica and G{\"o}berndorfer, Lisa and J{\"a}ger, Georg},
  journal={Heliyon},
  volume={10},
  number={22},
  year={2024},
  publisher={Elsevier}
}

@article{toleikyte2024clean,
  title={Clean Energy Technology Observatory: Heat Pumps in the European Union-2024 Status Report on Technology Development, Trends, Value Chains and Markets},
  author={Toleikyte, Agne and Lecomte, Eric and Volt, Jonathan and Lyons, Lorcan and Roca, Reina Juan Carlos and Georgakaki, Aliki and Letout, Simon and Mountraki, Aikaterini and Wegener, Moritz and Schmitz, Andreas and others},
  year={2024}
}

\end{document}